\documentclass[journal]{IEEEtran}
\usepackage{graphicx}
\usepackage{amsmath}
\usepackage{algorithm}
\usepackage{algorithmicx}
\usepackage{algpseudocode}
\ifCLASSINFOpdf
\else
\fi
\begin{document}
%
\title{Sequential Gating Ensemble Network  for Noise Robust Multi-Scale Face Restoration}
%
%
%

\author{Zhibo Chen,~\IEEEmembership{Senior~Member,~IEEE},
        ~Jianxin Lin, ~Tiankuang Zhou,
        ~and Feng Wu,~\IEEEmembership{Fellow,~IEEE}
\thanks{Zhibo Chen, Jianxin Lin, Tiankuang Zhou and Feng Wu are with University of Science and Technology of China, Hefei, Anhui, 230026, China, (e-mail: chenzhibo@ustc.edu.cn)}
\thanks{This work was supported in part by the National Key Research and Development Program of China under Grant No. 2016YFC0801001, NSFC under Grant 61571413, 61632001,61390514, and Intel ICRI MNC.}
\thanks{}}
%
%

\markboth{IEEE Transaction on Cybernetics Submission}%
{Shell \MakeLowercase{\textit{et al.}}: Sequential Gating Ensemble Network for Noise Robust Multi-Scale Face Restoration}
%



\maketitle

\begin{abstract}
Face restoration from low resolution and noise is important for applications of face analysis recognition. However, most existing face restoration models omit the multiple scale issues in face restoration problem, which is still not well-solved in research area. In this paper, we propose a Sequential Gating Ensemble Network (SGEN) for multi-scale noise robust face restoration issue. To endow the network with multi-scale representation ability, we first employ the principle of ensemble learning for SGEN network architecture designing. The SGEN aggregates multi-level base-encoders and base-decoders into the network, which enables the network to contain multiple scales of receptive field. Instead of combining these base-en/decoders directly with non-sequential operations, the SGEN takes base-en/decoders from different levels as sequential data. Specifically, it is visualized that SGEN learns to sequentially extract high level information from base-encoders in bottom-up manner and restore low level information from base-decoders in top-down manner. Besides, we propose to realize bottom-up and top-down information combination and selection with Sequential Gating Unit (SGU). The SGU sequentially takes information from two different levels as inputs and decides the output based on one active input. Experiment results on benchmark dataset demonstrate that our SGEN is more effective at multi-scale human face restoration with more image details and less noise than state-of-the-art image restoration models. Further utilizing adversarial training scheme, SGEN also produces more visually preferred results than other models under subjective evaluation.
\end{abstract}

\section{Introduction}\label{intro}
Facial analysis techniques, such as face recognition and face detection, have been widely studied and explored in the past decades. Meanwhile, to ensure the public security and accelerate the crime detection, the intelligent surveillance systems have been rapidly developed. Therefore, facial analysis techniques have been employed to various applications with surveillance systems, such as criminal investigation. However, the performance of most facial analysis techniques would degrade rapidly when given low quality face images. In real surveillance systems, the quality of the surveillance face images is affected by many factors, such as long distance between camera and object, and insufficient light in natural scene, which resulting in low resolution, noise and so on. Therefore, how to restore a high quality face from a low quality face is challenging. Face restoration technique provides a viable way to improve performance of facial analysis techniques on low quality face images.

\begin{figure}
  \centerline{\includegraphics[width=9.0cm]{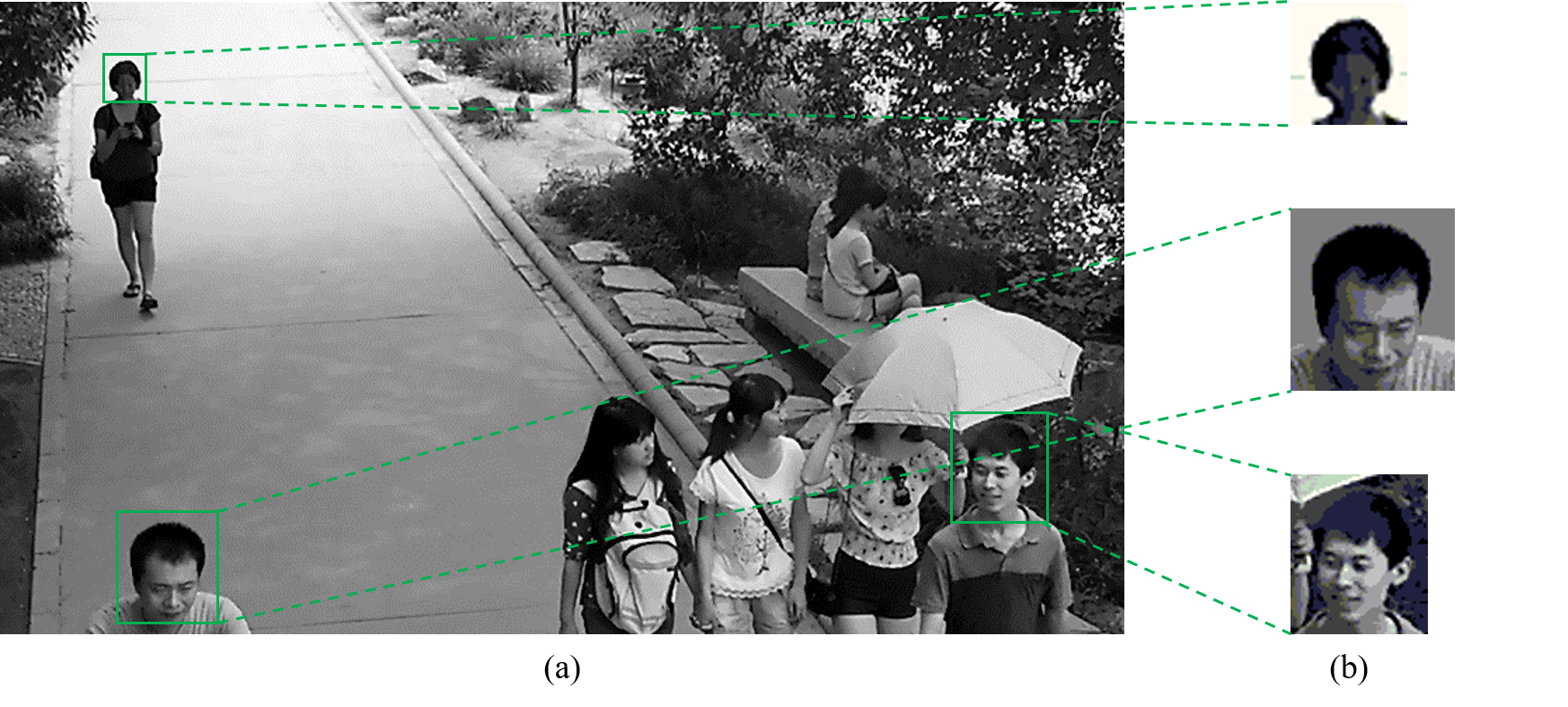}}
  \caption{Illustration of multi-scale face in one frame of surveillance video. (a) is the surveillance image from a surveillance camera; (b) are the three multi-scale face images extracted from (a).}
  \centering
\label{fig:fig3}
\end{figure}

Obtaining face images with abundant facial feature details is important for face restoration techniques, so numerous face restoration algorithms have been proposed in recent years. Some algorithms focus on solving face restoration from low-resolution (LR) problem (i.e., face hallucination), such as works in \cite{wang2005hallucinating,ma2010hallucinating,wang2014face,yu2016ultra, huang2011fast, zeng2018expanding}. To be consistent with more realistic situation, other algorithms also take the noise corruption into consideration during face super resolution, such as works in \cite{jiang2014noise,jiang2016noise}. We observe that most existing face restoration methods omit one vital characteristic of real-world images, namely images in real applications always contain faces of different scales as illustrated in Figure \ref{fig:fig3}. Also, when the images are corrupted with serious distortions, it's hard to extract the faces in the distorted images for face restoration since face detection methods may not work well under this situation. Therefore, in this paper, we focus on solving multi-scale face restoration close to real-world situation. The target of our proposed model is to effectively restore face images with details from noise corrupted LR face images without scale limitation.

Face restoration, which transfers low quality face image to high quality face image, can be considered as one of the image-to-image translation problem that transfers one image domain to another image domain. Solutions on image-to-image translation problem \cite{taigman2016unsupervised,yoo2016pixel,johnson2016perceptual} usually use autoencoder network \cite{hinton2006reducing} as a generator. However, single autoencoder network is too simple to represent multi-scale image-to-image translation due to lack of multi-scale representation. Meanwhile, ensemble learning, a machine learning paradigm where multiple learners are trained to solve the same problem, have shown its ability to make accurate prediction from multiple ``weak learners'' in classification problem \cite{kuncheva2004combining,polikar2006ensemble}. Therefore, an effective way to reinforce predictive performance of autoencoder network can be aggregating multiple base-generators into an enhanced-generator. In our model, we introduce base-encoders and base-decoders from low level to high level. These multi-level base-en/decoders ensure the generator have more diverse representation capacity to deal with multi-scale face image restoration.

The typical way of ensemble is to train a set of alternative models and takes a vote for these models. However, multi-scale face restoration is a problem that concerns multiple processes of feature abstracting and generating, merely taking a vote (or with other ensemble method) fails to incorporate high level information and restore detail information. Based on this observation, we devise a sequential ensemble structure that takes base-en/decoders from different levels as sequential data. The different combination directions of base-en/decoders are determined by the different goals of encoder and decoder. This sequential ensemble method is inspired by long short-term memory (LSTM) \cite{hochreiter1997long}. LSTM has been proved successfully in modeling sequential data, such as text and speech \cite{sundermeyer2012lstm,sutskever2014sequence}. The LSTM has the ability to optionally choose information passing through because of the gate mechanism. Specially, we design a Sequential Gating Unit (SGU) to realize information combination and selection, which sequentially takes base-en/decoders' information from two different levels as inputs and decides the output based on one active input.

Restoring low quality face image to high quality face image is an ill-posed problem, for which face details are usually absent in restored face images. Traditional optimization target of image restoration problem is to minimize the mean square error (MSE) between the restored image and the ground truth. However, minimizing MSE will often encourage smoothness in the restored image. Recently, generative adversarial networks (GANs) \cite{goodfellow2014generative,denton2015deep,radford2015unsupervised,salimans2016improved} show state-of-the-art performance on generating pictures of both high resolution and good semantic meaning. The high level real/fake decision made by discriminator causes the generated images' distribution close to the class of target domain, which endow the generated images with more details as target domain. Therefore, we utilize the adversarial learning process proposed in GAN \cite{goodfellow2014generative} for restoration model training.

In general, we propose to solve multi-scale face restoration problem with a Sequential Gating Ensemble Network (SGEN). The contribution of our approach includes three aspects:
\begin{itemize}
  \item We employ the principle of ensemble learning into network architecture designing. The SGEN is composed of multi-level base-en/decoders, which has better representation ability than ordinary autoencoder.
  \item The SGEN combines base-en/decoders from different levels with bottom-up and top-down manners corresponding to the different goals of encoder and decoder, which enables network to learn more compact high level information and restore more low level details.
  \item Furthermore, we propose a SGU unit to sequentially guide the information combination and selection from different levels.
\end{itemize}
The rest of this paper is organized as follows. We introduce related work in Section \ref{sec2} and present the details of proposed SGEN in Section \ref{sec3}, including network architecture, SGU unit and adversarial learning for SGEN. We present experiment results in Section \ref{sec4} and conclude in Section \ref{sec5}.
\section{Related Work}\label{sec2}
\begin{figure*}[htb!]
  \centerline{\includegraphics[width=17.5cm]{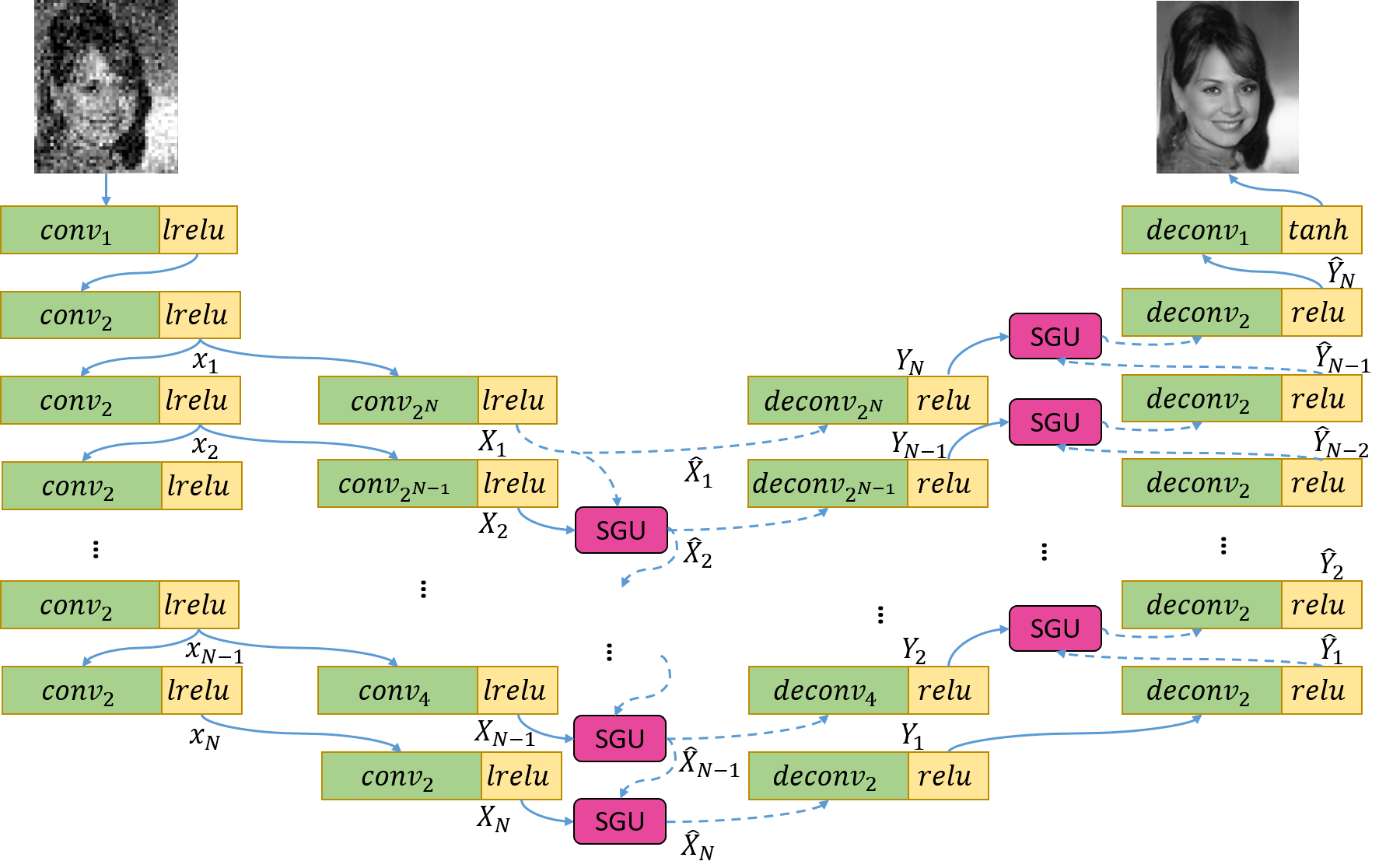}}
  \caption{Sequential ensemble network architecture of SGEN. Convolution and pooling operations are shown in green, activation functions are shown in yellow and the SGU is shown in pink.}
  \centering
\label{fig:fig1}
\end{figure*}
face restoration is of great importance for vision applications. Therefore, extensive studies have been carried out to restore the low quality face image to high quality face image in the past decades. The early face restoration algorithms can be categorized into two classes, i.e., global face-based restoration methods and local patch-based restoration methods. Global face-based restoration methods model LR face image as a linear combination of LR face images in the training set by using different face representation models, such as principal component analysis (PCA) \cite{wang2005hallucinating}, kernel PCA \cite{chakrabarti2007super}, locality preserving projections \cite{zhuang2007hallucinating}, canonical correlation analysis (CCA) \cite{huang2010super}, and non-negative matrix factorization \cite{yang2010image}. Then, these global face-based restoration methods reconstruct the target HR face image by replacing the LR training images with the corresponding HR ones, while using the same coefficients. Though global face-based restoration methods may well preserve the global shape information, the details of input face are usually not well recovered by these methods.

To overcome the drawback of global face-based restoration methods, local patch-based restoration methods decompose face image into small patches, which can capture more facial details. Local patch-based restoration methods assume that LR and HR face patch manifolds are locally isometric. Therefore, once obtaining the representation of the input LR patch with the LR training patches, we can reconstruct the target HR patch by transforming the reconstruction weights to corresponding HR training patch. The work in \cite{ma2010hallucinating} proposed a least squares representation (LSR) framework that restores images using all the training patches, which incorporates more face priors. Due to the un-stability of LSR, \cite{wang2014face} introduced a weighted sparse representation (SR) with sparsity constraint for face super-resolution. However, one main drawback of SR based methods is its sensitivity to noise. Accordingly, \cite{jiang2014noise,jiang2016noise} introduced to reconstruct noise corrupted LR images with weighted local patch, namely locality-constrained representation (LcR).

In the past few years, convolutional neural networks (CNN) \cite{lecun1989backpropagation} have shown an explosive popularity and success in various computer vision fields, such as image recognition \cite{he2016deep}, object detection \cite{ren2015faster}, face recognition \cite{parkhi2015deep}, and semantic segmentation \cite{long2015fully}. CNN based image restoration algorithms have also shown excellent performance compared with previous state-of-the-art methods. SRCNN \cite{dong2014learning} is a three layer fully convolutional network and trained end-to-end for image super resolution. \cite{yu2016ultra} presented a ultra-resolution discriminative generative network (URDGN) that can ultra-resolve a very low resolution face. Instead of building network as simple hierarchy structure, other works also applied the skip connections, which can be viewed as one kind of ensemble structure \cite{veit2016residual}, to image restoration tasks. \cite{ledig2017photo} proposed a SRResNet that uses ResNet blocks in the generative model and achieves state-of-the-art peak signal-to-noise ratio (PSNR) performance for image super-resolution. In addition, they presented a SRGAN that utilizes adversarial training to achieve better visual quality than SRResNet. \cite{mao2016image} proposed a residual encoder-decoder network (RED-Net) which symmetrically links convolutional and deconvolutional layers with skip-layer connections.

However, these skip-connections in \cite{ledig2017photo,mao2016image} fail to explore the underlying sequential relationship among multi-level feature maps in image restoration problem. Therefore, we design our SGEN followed by the goal of autoencoder, which sequentially extracts high level information from base-encoders in bottom-up manner and restores low level information from base-decoders in top-down manner.

\section{Sequential Gating Ensemble Network}\label{sec3}
Architecture of our Sequential Gating Ensemble Network (SGEN) is shown in Figure \ref{fig:fig1}. We discussed the details of SGEN in the following subsections. Firstly, we introduce the sequential ensemble network architecture of SGEN. Then we present the Sequential Gating Unit (SGU) for combining the multi-level information. Finally, we elaborate the adversarial training for SGEN and the loss function for adversarial training process.

\subsection{Sequential ensemble network architecture}\label{sec3.1}
First, our generator is a fully convolutional computation network \cite{long2015fully} that can take arbitrary-size inputs and predict dense outputs. Then, let us denote $k$-$th$ encoder feature, $k$-$th$ base-encoder feature, $k$-$th$ combined base-encoder feature, $k$-$th$ base-decoder feature,  $k$-$th$ combined base-decoder feature by $x_{k}$, $X_{k}$, $\hat{X}_k$, $Y_{k}$, $\hat{Y}_k$ respectively, and suppose there are $N$ base-encoders and base-decoders in total. Given a low quality face image sample $s$, the SGEN $G$ in Figure \ref{fig:fig1} can be illustrated in the formulas below:
\begin{equation}\label{7}
{x_1} = lrelu(con{v_2}(lrelu(conv_1(s)))),
\end{equation}
\begin{equation}\label{8}
{x_k} = lrelu(con{v_2}({x_{k - 1}})), \quad k = 2,3,...,N
\end{equation}
\begin{equation}\label{9}
{X_k} = lrelu(con{v_{{2^{N-k+1}}}}({x_k})), \quad k = 1,2,...,N
\end{equation}
\begin{equation}\label{10-1}
{\hat{X}_1} = X_1,
\end{equation}
\begin{equation}\label{10}
{\hat{X}_k} = SGU(X_k, \hat{X}_{k-1}), \quad k = 2,3,...,N
\end{equation}

\begin{equation}\label{11}
{Y_k} = relu(decon{v_{2^{k}}}(\hat{X}_{N-k+1})), \quad k = 1,2,3,...,N
\end{equation}
\begin{equation}\label{12}
{\hat{Y}_1} = relu(decon{v_{2}}(Y_1)),
\end{equation}
\begin{equation}\label{13}
{\hat{Y}_k} = relu(decon{v_{2}}(SGU(Y_k, \hat{Y}_{k-1})), \quad k = 2,3,...,N
\end{equation}
\begin{equation}\label{14}
G(s) = tanh(conv_1(\hat{Y}_N)),
\end{equation}
where $G(s)$ is the generated face image, $con{v_{2^k}}$ and $decon{v_{2^k}}$ are convolution and de-convolution operations with factor $2^k$ pooling and upsampling respectively. SGU is sequential gating unit. Each de-convolution layer is followed by $relu$ (rectified linear unit) \cite{nair2010rectified}, and each convolution layer is followed by $lrelu$ (leaky relu) \cite{maas2013rectifier}, except for the last layer of generator (using $tanh$ activation function). Note, there is no parameters sharing among different convolution $conv$ operations, de-convolution $deconv$ operations and SGUs in this paper.

The bottom-up base-encoders combination and top-down base-decoders combination are determined by the different goals of encoder and decoder. Given a low quality face image input, the encoder of a autoencoder would like to transfer the input into highly compact representation with semantic meaning (i.e., bottom-up information extraction), and the decoder would like to restore the face image with abundant details (i.e., top-down information restoration). Therefore, without breaking the rules of autoencoder, we combine the multi-level base-en/decoders in two directions. Accordingly, we design a SGU to realize the multi-level information combination and selection in en/decoder stage.

Combination of these multi-level base-en/decoders provides another benefit that network layer of SGEN contains multiple scales of receptive field, which helps the encoder learn features with multi-scale information and helps decoder generate more accurate images from multi-scale features. Experiment results also demonstrate that our network is more capable of restoring multi-scale low quality face image than other networks.
\subsection{Sequential gating unit}\label{sec3.2}
\begin{figure}
  \centerline{\includegraphics[width=8.5cm]{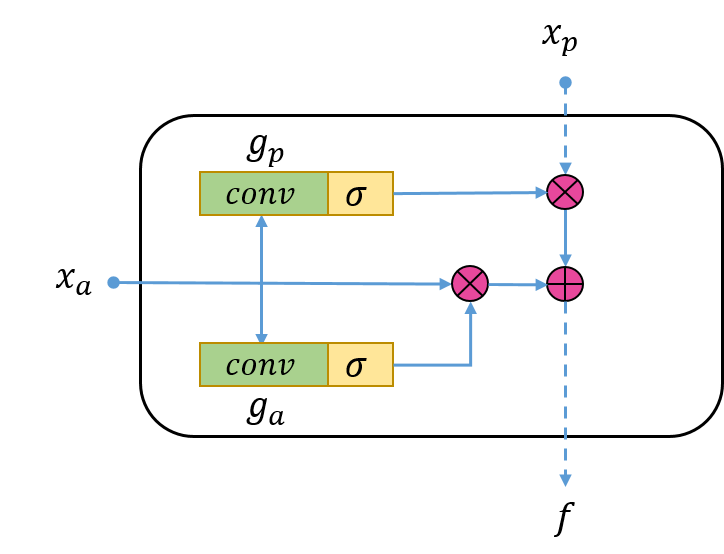}}
  \caption{Sequential Gating Unit. Element-wise multiplications and additions are shown in pink.}
  \centering
\label{fig:fig2}
\end{figure}

To further utilize the sequential relationship among multi-level base-en/decoders, we propose a Sequential Gating Unit (SGU) to sequentially combine and select the multi-level information, which takes base-en/decoders' information from two different levels as inputs and decides the output based on one active input. The SGU is shown in the Figure \ref{fig:fig2}, equation depicting the unit is given as below:
\begin{equation}\label{15}
\begin{split}
{f} &= SGU(x_a,x_p)\\ &= g_a(x_a)\cdot x_a+g_p(x_a)\cdot x_p,
\end{split}
\end{equation}
where $f$ is the SGU output, $g_a$ and $g_p$ are two non-linear transform ``gate'', $\sigma(.)$ in Figure \ref{fig:fig2} is sigmoid activation function ($0 \le \sigma(.)\le 1$), $x_a$ and $x_p$ are active input and passive input respectively. Although both $g_a$ and $g_p$ consist of one convolution layer and one sigmoid activation function, they do not have any shared parameters. The active input $x_a$ makes the decision what information we are going to throw away from the passive input $x_p$ and what new information we are going to add from the active input itself. In particular, observe that if $g_a(x_a)=\mathbf{1}$ and $g_p(x_a)=\mathbf{0}$, the active input $x_a$ throw away all the information from the passive input $x_p$. Conversely, if $g_a(x_a)=\mathbf{0}$ and $g_p(x_a)=\mathbf{1}$, the active input $x_a$ choose to pass all the information from the passive input $x_p$ rather than itself. Therefore, SGU can smoothly vary its behavior on information combination and selection, and is optimized by the whole training objective.

In the encoder stage, high level base-encoder acts as $x_a$ and takes control over the low level information, which sequentially updates the high level semantic information and removes noise. For the decoder stage, the low level base-decoder becomes $x_a$ and takes control over the high level information in an opposite direction, which sequentially restores low level information and generates images with more details.

\subsection{Training Algorithm}\label{sec3.3}
We apply the adversarial training of GAN in our proposed model. The adversarial training needs to learn a discriminator $D$ to guide the generator $G$, i.e. SGEN in this paper, to produce realistic target under the supervision of real/fake. In face restoration case, the objective function of GAN can be represented as minimax function:
\begin{equation}\label{16}
\begin{split}
\mathop {\min }\limits_G \mathop {\max }\limits_D \ell_{\text{GAN}}(s,t) &= {E_{t \sim {p_\mathcal{T}}(t)}}[\log (D(t))]\\ &+{E_{s \sim {p_\mathcal{S}}(s)}}[\log (1 - D(G(s))],
\end{split}
\end{equation}
where $s$ is a sample from the low quality source domain $\mathcal{S}$ and $t$ is the corresponding sample in high quality target domain $\mathcal{T}$. In addition to using adversarial loss in the generator training process, we add the mean square error (MSE) loss for generator to require the generated image $G(s)$ as close as to the ground truth value of pixels. The modified loss function for adversarial SGEN training is shown as below:
\begin{equation}\label{17}
\begin{split}
\mathop {\min }\limits_G \mathop {\max }\limits_D \ell_{\text{GAN}}(s,t) &= {E_{t \sim {p_\mathcal{T}}(t)}}[\log (D(t))]\\&+{E_{s \sim {p_\mathcal{S}}(s)}}[\log (1 - D(G(s))]\\&+\lambda {\ell_{\text{MSE}}}(s,t),
\end{split}
\end{equation}
\begin{equation}\label{18}
{\ell_{\text{MSE}}}(s,t)={E_{s \sim {p_\mathcal{S}}(s),t \sim {p_\mathcal{T}}(t)}}{[||t - G(s)|{|_2^2}]},
\end{equation}
where $\lambda$ is weight to achieve balance between adversarial term and MSE term.

To make the discriminator be able to take input of arbitrary size as well, we design a fully convolutional discriminator with global average pooling proposed in \cite{lin2013network}. We replace the traditional fully connected layer with global average pooling. The idea of global average pooling is to take the average of each feature map as the resulting vector fed into classification layer. Therefore, the discriminator has much fewer network parameters than fully connected network and overfitting is more likely to be avoided.

We summarize the training process in Algorithm~\ref{alg_1}. In Algorithm~\ref{alg_1}, the choice of optimizers $Opt(\cdot,\cdot)$ is quite flexible, whose two inputs are the parameters to be optimized and the corresponding gradients. One can choose different optimizers (e.g. Adam~\cite{kingma2014adam}, or nesterov gradient descend~\cite{nesterov1983method}) for different tasks, depending on common practice for specific tasks and personal preferences. Besides, the $G$ and $D$ might refer to either the models themselves, or their parameters, depending on the context.
\begin{algorithm}
\caption{SGEN training process}
\label{alg_1}
\begin{algorithmic}[1]
\Require Training images $\{s_{i}\}_{i=1}^{m}\subset\mathcal{S} $, $\{t_{j}\}_{j=1}^{m}\subset\mathcal{T}$, batch size $K$, optimizer $Opt(\cdot,\cdot)$;
\State Randomly initialize $G$ and $D$.
\State Randomly sample a minibatch of images and prepare the data pairs  $\mathcal{P}=\{(s_{k},t_{k})\}^K_{k=1}$.
\State For any data pair $(s_{k},t_{k})\in\mathcal{P}$, generate reconstructed images by Eqn.(\ref{7}-\ref{14}).
\State Update the discriminators as follows:\newline
$D \leftarrow Opt(D, (1/K)\nabla_{D}\textstyle{\sum_{k=1}^{K}}-\ell_{\text{GAN}}(s_k,t_k))$.
\State Update the SGEN, i.e., $G$ as follows:\newline
$G \leftarrow Opt(G, (1/K)\nabla_{G}\textstyle{\sum_{k=1}^{K}}\ell_{\text{GAN}}(s_k,t_k))$.
\State Repeat step 2 to step 5 until convergence
\end{algorithmic}
\end{algorithm}
\subsection{Discussion}
According to the network architecture described in Section \ref{sec3.1} and Section \ref{sec3.2}, in the encoder stage, high level base-encode acts as $x_a$ and takes control over the low level information $x_p$, while in the decoder stage, the low level base-decoder becomes $x_a$ and takes control over the high level information in an opposite direction, this kind of sequential ensemble process can be denoted as below:
\begin{equation}\label{10_B}
\begin{split}
{\hat{X}_k} &= SGU(X_k, \hat{X}_{k-1})\\&= g_a(X_k)\cdot X_k+g_p(X_k)\cdot \hat{X}_{k-1}\\&=g_a(X_k)\cdot X_k+g_p(X_k)\cdot \\ &(g_a(X_{k-1})\cdot X_{k-1}+g_p(X_{k-1})\cdot \hat{X}_{k-2}) \\  &\quad \quad \quad \vdots \quad, \\ k& = 2,3,...,N.
\end{split}
\end{equation}
\begin{equation}\label{13_B}
\begin{split}
{\hat{Y}_k} &= relu(decon{v_{2}}(SGU(Y_k, \hat{Y}_{k-1}))\\ &= relu(decon{v_{2}}(g_a(Y_k)\cdot Y_k+g_p(Y_k)\cdot \hat{Y}_{k-1}))\\ &= relu(decon{v_{2}}(g_a(Y_k)\cdot Y_k+g_p(Y_k) \\ &\cdot (relu(decon{v_{2}}(g_a(Y_{k-1})\cdot Y_{k-1}+g_p(Y_{k-1})\cdot \hat{Y}_{k-2}))))) \\ &\quad \quad \quad \vdots \quad, \\k& = 2,3,...,N.
\end{split}
\end{equation}
As we can see from Eqn.(\ref{10_B}), the high level base-encodes sequentially choose to updates the high level semantic information following a bottom-up information flow, which effectively cleans the corrupted input and removes noise during this process. Similarly, in Eqn.(\ref{13_B}), the low level base-decoder sequentially restores low level information following a top-down information flow, which effectively reconstructs images with more face details. We also visualize the gates in the SGU in the experiments section \ref{exp_vis} and further verify the effectiveness of information selection by using SGU.

In particular, we observe that if we set $g_a(X_k)=\mathbf{1}$, $g_p(X_k)=\mathbf{0}$, $g_a(Y_k)=\mathbf{1}$ and $g_p(Y_k)=\mathbf{1}$, we have ${\hat{X}_k} = X_k$ and ${\hat{Y}_k} = relu(decon{v_{2}}(Y_k+\hat{Y}_{k-1})) = relu(decon{v_{2}}(relu(decon{v_{2^{k}}}(\hat{X}_{N-k+1}))+\hat{Y}_{k-1}))$. Under such condition, the convolutional feature maps $\hat{X}_{N-k+1}=X_{N-k+1}$ are directly passed to the decoder, and summed to deconvolutional feature maps $\hat{Y}_{k-1}$ after one deconvolutional layer, which basically imitates the skip-connections in RED-Net \cite{mao2016image}. Also, if we set $g_a(X_k)=\mathbf{1}$, $g_p(X_k)=\mathbf{1}$, $g_a(Y_k)=\mathbf{1}$ and $g_p(Y_k)=\mathbf{1}$, we have ${\hat{X}_k} = X_k+\hat{X}_{k-1}$ and ${\hat{Y}_k} = relu(decon{v_{2}}(Y_k+\hat{Y}_{k-1}))$. In this case, our SGEN can provide extra residual connection in the encoder stage compared with RED-Net, which will be more likely to avoid gradient vanishing and be easier to train as explained in \cite{he2016deep}. Thus, depending on the output of the gates in SGU, our SGEN can smoothly vary its behavior between a plain network and residual networks.
\section{Experiments}\label{sec4}
\subsection{Parameters setting}
\begin{figure*}
  \centerline{\includegraphics[width=17.5cm]{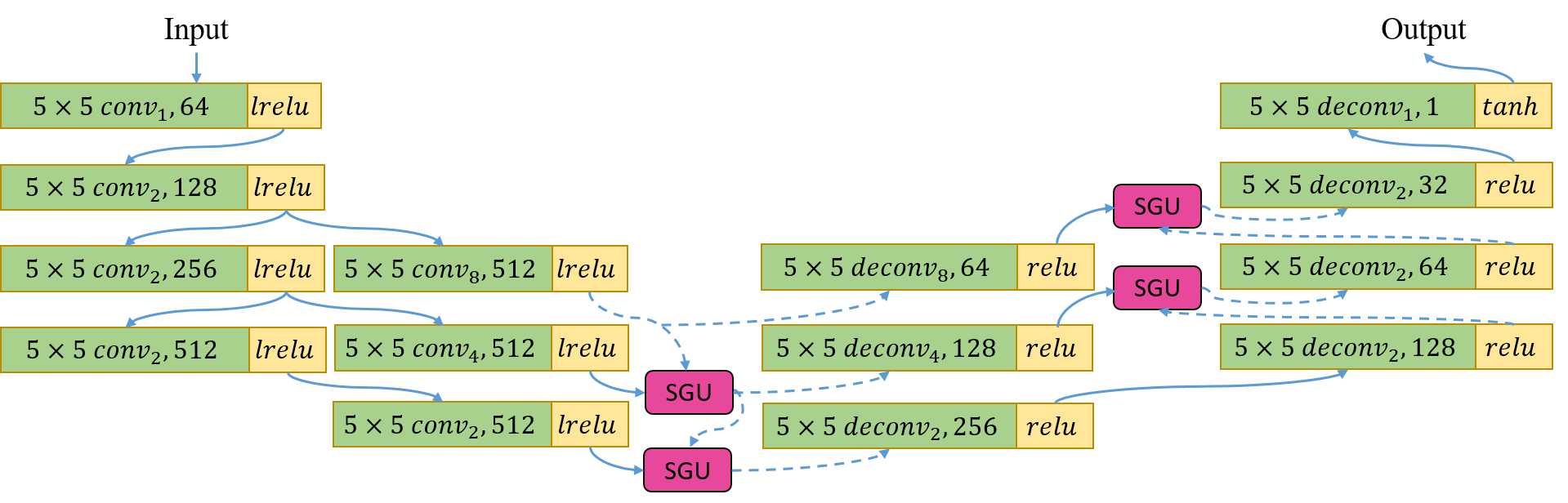}}
  \caption{Network configuration of SGEN ($N=3$). A convolution layer is represented as {\emph{\{kernel size\}$\!\times\!$\{kernel size\} $conv_{2^k}$, \{channel number\}}}. A deconvolution layer is represented as {\emph{\{kernel size\}$\!\times\!$\{kernel size\} $deconv_{2^k}$, \{channel number\}}}.}
  \centering
\label{fig:fig4}
\end{figure*}

In the following experiments, our SGEN's network configuration ($N=3$) is shown in Figure \ref{fig:fig4}. We set $N=3$ levels for the SGEN to achieve a trade-off between performance and computation cost. SGU contains two convolution modules whose filter number is same to the inputs' channel number. The discriminator architecture consists of $4$ convolution layers followed by $lrelu$ units, then a global average pooling layer with $512$ and a fully connected layer with $1$ neurons follows. The final sigmoid unit produces the probability of a sample coming from real data. We use the adaptive learning method Adam \cite{kingma2014adam} as the optimization algorithm with learning rate of $0.0001$. Minibatch size is set to $64$ for every experiments. The weight $\lambda$ is set to $10$. An
implementation of SGEN is available in https://github.com/tomorrowi6/SGEN-pytorch.git.
\subsection{Dataset and evaluation metrics}
We carry out experiments below on the widely used face dataset CelebA \cite{liu2015faceattributes} containing $202599$ cropped celebrity faces. We set aside $20000$ images as test set, $20000$ images as validation set, and the rest as training set. We resize the face images from $128 \times 96$ to $208 \times 176$ which are commonly used resolutions in practice. Specially, we sample $6$ different scales between two resolutions. Then we downsample the images to low-resolution (LR) by a factor of $4$ and corrupt the images by AWGN noise (standard deviation $\delta = 30$). To enable the noise corrupted LR images as the network input, we up-sample the LR images by a factor $4$ using nearest interpolation.

Restoration results are evaluated with peak signal-to-noise ratio (PSNR) and structure similarity index (SSIM) \cite{wang2004image}. However, PSNR and SSIM are objective metrics that can not always be consistent with human perceptual quality. Therefore, we conduct subjective evaluation for restoration results to further investigate the effectiveness of our model. We recruited $18$ subjects for subjective evaluation. We show each subject the corresponding source images prior to their evaluation for the generated images. Thus, they can form a general quality standard of generated images. After viewing one test image, subject gives a quality score from $1$ (bad quality) to $5$ (excellent quality). For each model, $288$ restoration images from six different scales are evaluated, and mean opinion scores (MOS) are computed at each scale.

\subsection{Comparison of different losses and ensemble methods}
\begin{figure}
  \centerline{\includegraphics[width=8.5cm]{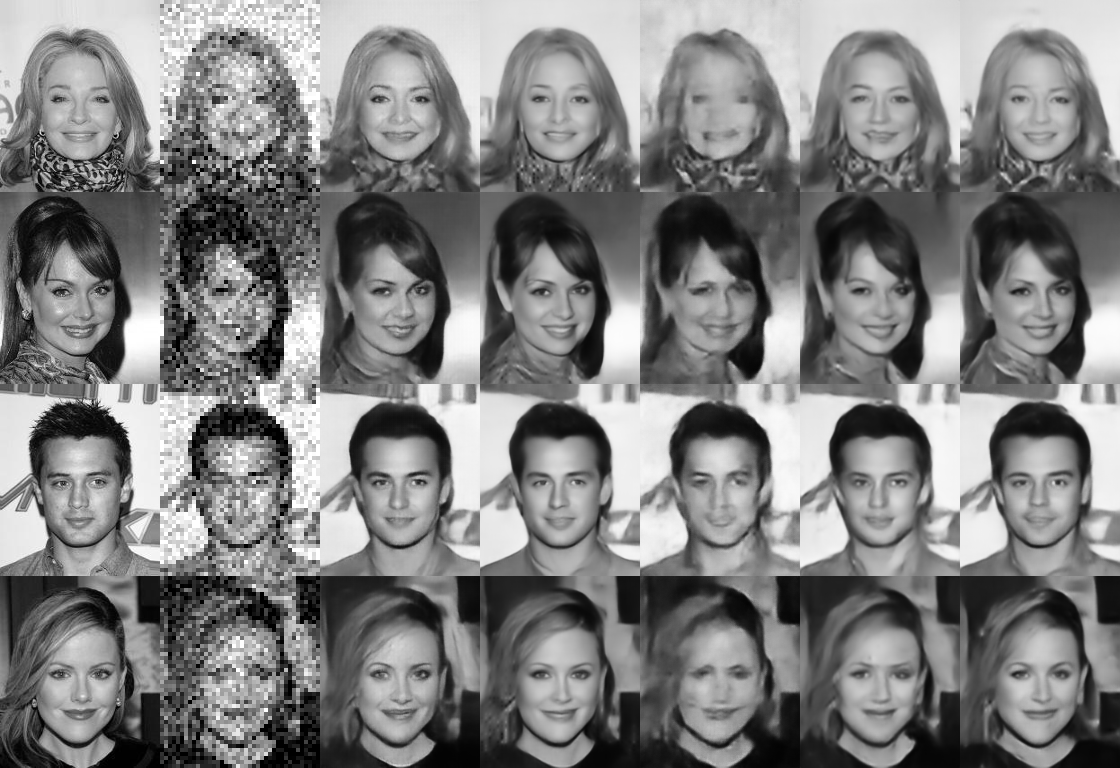}}
  \caption{Restoration samples from the same scale. From left to right: ground truth, noise corrupted LR images, results from SGEN, results from SGEN-MSE, results from MEN, results from AEN, results from CEN.}
  \centering
\label{fig:fig5}
\end{figure}
\begin{figure}
  \centerline{\includegraphics[width=8.5cm]{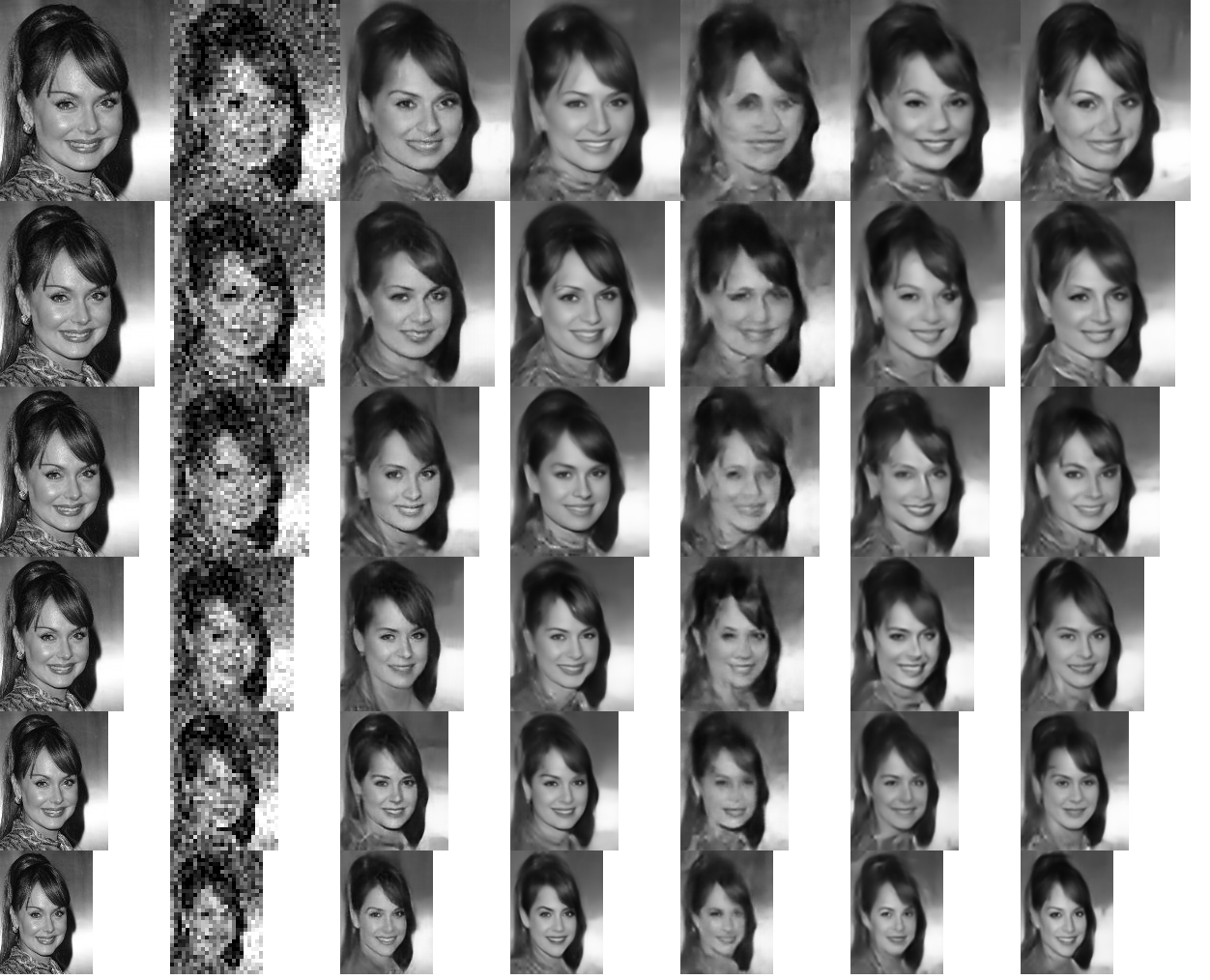}}
  \caption{Restoration samples from the six different scales. From left to right: ground truth, noise corrupted LR images, results from SGEN, results from SGEN-MSE, results from MEN, results from AEN, results from CEN.}
  \centering
\label{fig:fig7}
\end{figure}

\begin{table*}[]
\centering
\caption{Average PSNR, SSIM and MOS results of networks with different loss and ensemble methods from scale $128 \times 96$ to scale $160 \times 128$. Highest scores are in bold.}
\label{table3}
\begin{tabular}{|c|c|c|c|c|c|c|c|c|c|}
\hline
                  & \multicolumn{3}{c|}{Scale $128 \times 96$}                    & \multicolumn{3}{c|}{Scale $144 \times 112$}                   & \multicolumn{3}{c|}{Scale $160 \times 128$}                     \\ \hline
                  & \textbf{PSNR}  & \textbf{SSIM}   & \textbf{MOS} & \textbf{PSNR}  & \textbf{SSIM}  & \textbf{MOS} & \textbf{PSNR}  & \textbf{SSIM}   & \textbf{MOS} \\ \hline
\textbf{SGEN}     & 22.37         & 0.6555          &\textbf{4.4833}              & 23.08         & 0.6863          &\textbf{4.7833}              & 23.61          & 0.7006          &\textbf{4.6333}
              \\ \hline
\textbf{SGEN-MSE} & \textbf{23.00}   & \textbf{0.6989} &4.3667    & \textbf{23.60} & \textbf{0.7161} &4.5000    & \textbf{24.12} & \textbf{0.7327} &4.6000              \\ \hline
\textbf{MEN}      & 22.10          & 0.6485          &2.0833              & 22.68         & 0.6652          &2.1333              & 23.04          & 0.6807          &2.0833              \\ \hline
\textbf{AEN}      & 22.61         & 0.6800            &3.2000              & 23.17         & 0.7015          &3.4000              & 23.64          & 0.7134          &3.2000              \\ \hline
\textbf{CEN}      & 22.75         & 0.6945          &4.0333              & 23.36         & 0.7129          &4.1000              & 23.88          & 0.7261          &4.2000              \\ \hline
\end{tabular}
\end{table*}
\begin{table*}[]
\centering
\caption{Average PSNR, SSIM and MOS results of networks with different loss and ensemble methods from scale $176 \times 144$ to scale $208 \times 176$. Highest scores are in bold.}
\label{table4}
\begin{tabular}{|c|c|c|c|c|c|c|c|c|c|}
\hline
                  & \multicolumn{3}{c|}{Scale $176 \times 144$}                    & \multicolumn{3}{c|}{Scale $192 \times 160$}                   & \multicolumn{3}{c|}{Scale $208 \times 176$}                   \\ \hline
                  & \textbf{PSNR}  & \textbf{SSIM}   & \textbf{MOS} & \textbf{PSNR}  & \textbf{SSIM}  & \textbf{MOS} & \textbf{PSNR} & \textbf{SSIM}   & \textbf{MOS} \\ \hline
\textbf{SGEN}     & 24.12          & 0.7202          &\textbf{4.6333}              & 24.55          & 0.733           &4.6000              & 24.92          & 0.7429          &\textbf{4.6833}              \\ \hline
\textbf{SGEN-MSE} & \textbf{24.61} & \textbf{0.7501} &4.3667    & \textbf{24.98} & \textbf{0.7576} & \textbf{4.6333}    & \textbf{25.39} & \textbf{0.7686} &4.5833              \\ \hline
\textbf{MEN}      & 23.50           & 0.6942          &1.9833              & 23.97          & 0.7052          &2.1167              & 24.46          & 0.7197          &2.2667              \\ \hline
\textbf{AEN}      & 24.14          & 0.7274          &3.2333              & 24.62          & 0.7411          &3.3000              & 24.97          & 0.7533          &3.2667              \\ \hline
\textbf{CEN}      & 24.42          & 0.7397          &4.1000              & 24.82          & 0.7545          &4.0500              & 25.20           & 0.7646          &4.0833              \\ \hline
\end{tabular}
\end{table*}

To investigate influence of different loss choices and verify the effectiveness of sequential ensemble structure of SGEN, we compare performance of following models:
\begin{itemize}
  \item \textbf{SGEN}. SGEN is trained with MSE and adversarial loss.
  \item \textbf{SGEN-MSE}. SGEN-MSE is trained with only MSE loss.
  \item \textbf{MEN}. MEN (Max Ensemble Network) is just like SGEN but without any SGU during the combination of base-en/decoders, it uses max ensemble instead.
  \item \textbf{AEN}. AEN (Average Ensemble Network) uses average ensemble instead of SGU.
  \item \textbf{CEN}. AEN (Concatenate Ensemble Network) uses concatenate ensemble instead of SGU.
\end{itemize}
The objective and subjective results are shown in Table \ref{table3} and Table \ref{table4}. Visual restoration samples from one scale and six scales are shown in Figure \ref{fig:fig5} and Figure \ref{fig:fig7}. Only MSE loss for SGEN training provides higher PSNR and SSIM than combining MSE and adversarial loss, this result is not surprising because only minimizing MSE is equal to maximize PSNR. Smoother restoration results are also obtained by SGEN-MSE and are less visually preferred than SGEN through MOS evaluation. Comparing SGEN with other non-sequential ensemble networks, SGEN achieves better subjective scores and produces visually preferred images with more face details. Actually, because of the gate mechanism in SGU, sequential ensemble method can be viewed as a generalized ensemble method including max ensemble and average ensemble. The CEN shows more competitive results than MEN and AEN because CEN combines base-en/decoders' information all together without any information selection. The better performance of SGEN than CEN also demonstrates the effectiveness of automatically information selection in sequential ensemble method.

\subsection{Comparison with state-of-the-art algorithms}\label{4.4}
\begin{table*}[]
\centering
\caption{Average PSNR, SSIM and MOS comparison results with state-of-the-art algorithms from scale $128 \times 96$ to scale $160 \times 128$. Highest scores are in bold.}
\label{table1}
\begin{tabular}{|c|c|c|c|c|c|c|c|c|c|}
\hline
                  & \multicolumn{3}{c|}{Scale $128 \times 96$}                    & \multicolumn{3}{c|}{Scale $144 \times 112$}                   & \multicolumn{3}{c|}{Scale $160 \times 128$}                     \\ \hline
                  & \textbf{PSNR}  & \textbf{SSIM}   & \textbf{MOS} & \textbf{PSNR}  & \textbf{SSIM}  & \textbf{MOS} & \textbf{PSNR}  & \textbf{SSIM}   & \textbf{MOS} \\ \hline
\textbf{SGEN}     & 22.37         & 0.6555          &\textbf{4.4833}              & 23.08         & 0.6863          &\textbf{4.7833}              & 23.61          & 0.7006          &\textbf{4.6333}
              \\ \hline
\textbf{SGEN-MSE} & \textbf{23.00}   & \textbf{0.6989} &4.3667    & \textbf{23.60} & \textbf{0.7161} &4.5000    & \textbf{24.12} & \textbf{0.7327} &4.6000              \\ \hline
SRCNN             & 21.72         & 0.5923          &1.0000              & 22.22         & 0.6094          &1.0667              & 22.69          & 0.6236          &1.1000              \\ \hline
SRResNet          & 22.73         & 0.6827          &3.0333              & 23.29         & 0.7016          &3.0667              & 23.81          & 0.7166          &3.2667              \\ \hline
SRGAN          & 22.29         & 0.6486          &2.9444              & 22.78         & 0.6796          &3.0000             & 23.43          & 0.6927          &3.3889              \\ \hline
RED-Net           & 22.77         & 0.6809          &3.6667              & 23.32         & 0.7001          &3.6333              & 23.83          & 0.7147          &3.8167              \\ \hline
URDGN             & 22.54         & 0.6688          &2.8667              & 23.10          & 0.6885          &3.0667              & 23.56          & 0.7044          &3.0500              \\ \hline
\end{tabular}
\end{table*}
\begin{table*}[]
\centering
\caption{Average PSNR, SSIM and MOS comparison results with state-of-the-art algorithms from scale $176 \times 144$ to scale $208 \times 176$. Highest scores are in bold.}
\label{table2}
\begin{tabular}{|c|c|c|c|c|c|c|c|c|c|}
\hline
                  & \multicolumn{3}{c|}{Scale $176 \times 144$}                    & \multicolumn{3}{c|}{Scale $192 \times 160$}                   & \multicolumn{3}{c|}{Scale $208 \times 176$}                   \\ \hline
                  & \textbf{PSNR}  & \textbf{SSIM}   & \textbf{MOS} & \textbf{PSNR}  & \textbf{SSIM}  & \textbf{MOS} & \textbf{PSNR} & \textbf{SSIM}   & \textbf{MOS} \\ \hline
\textbf{SGEN}     & 24.12          & 0.7202          &\textbf{4.6333}              & 24.55          & 0.7330           &4.6000              & 24.92          & 0.7429          &\textbf{4.6833}              \\ \hline
\textbf{SGEN-MSE} & \textbf{24.61} & \textbf{0.7501} &4.3667    & \textbf{24.98} & \textbf{0.7576} & \textbf{4.6333}    & \textbf{25.39} & \textbf{0.7686} &4.5833              \\ \hline
SRCNN             & 23.17          & 0.6419          &1.1500              & 23.58          & 0.6548          &1.2333              & 23.92          & 0.6643          &1.4833              \\ \hline
SRResNet          & 24.31          & 0.7328          &3.3500              & 24.77          & 0.7458          &3.0833              & 25.04          & 0.7543          &3.1500              \\ \hline
SRGAN          & 24.03          & 0.7198          &3.8333              & 24.37          & 0.7297          &3.9444              & 24.71          & 0.7413          &3.8889              \\ \hline
RED-Net           & 24.30           & 0.7318          &3.6667              & 24.73          & 0.7461          &3.6500              & 25.11          & 0.7560           &3.7167              \\ \hline
URDGN             & 23.98          & 0.7181          &2.9333              & 24.42          & 0.7306          &2.9500              & 24.67          & 0.7396          &2.8500              \\ \hline
\end{tabular}
\end{table*}
\begin{figure}
  \centerline{\includegraphics[width=8.5cm]{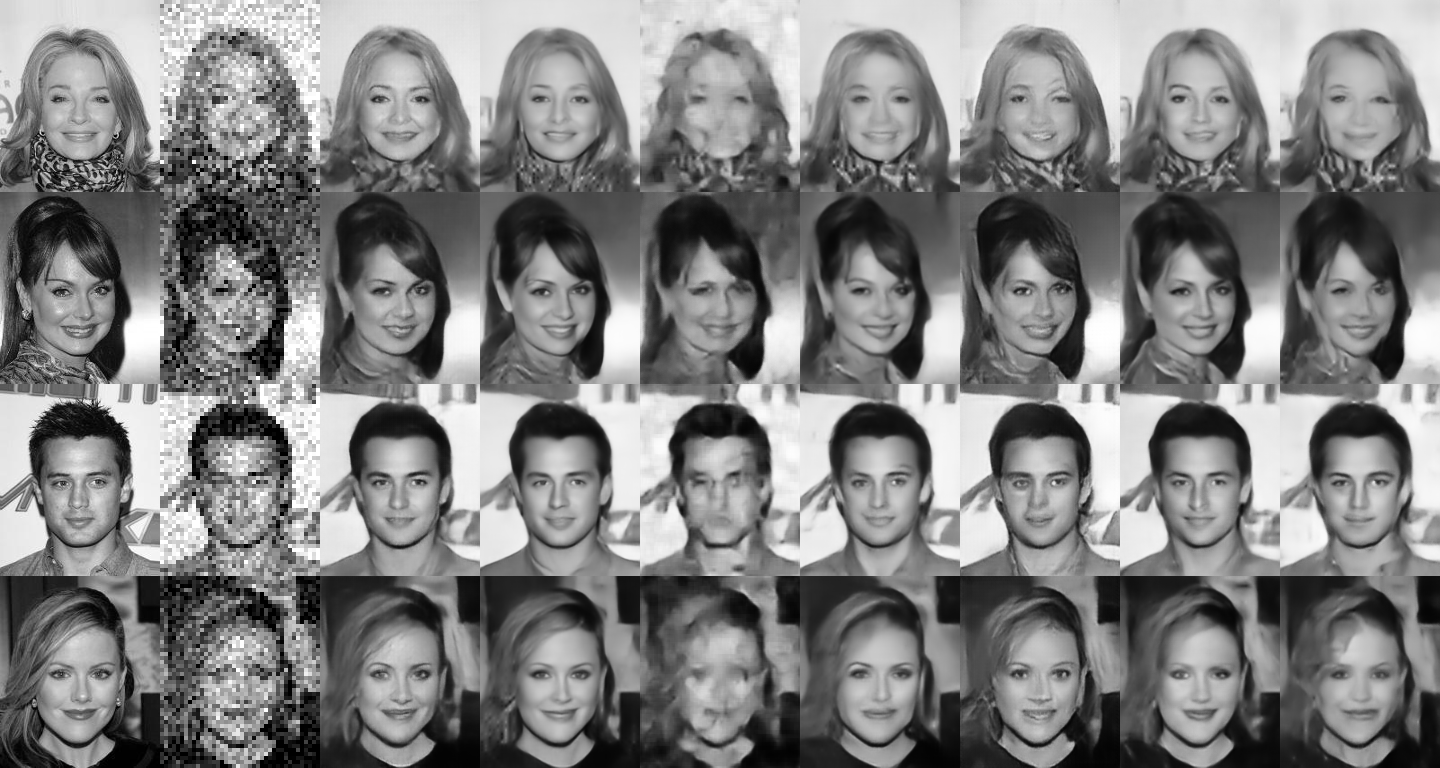}}
  \caption{Restoration samples from the same scale. From left to right: ground truth, noise corrupted LR images, results from SGEN, results from SGEN-MSE, results from SRCNN, results from SRResNet, results from SRGAN, results from RED-Net, results from URDGN.}
  \centering
\label{fig:fig6}
\end{figure}
\begin{figure}
  \centerline{\includegraphics[width=8.5cm]{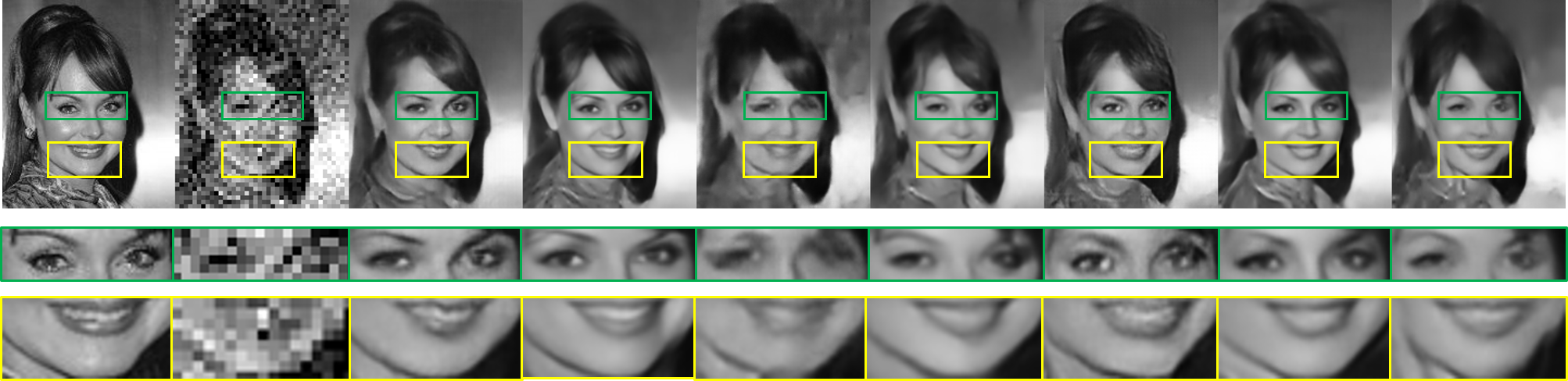}}
  \caption{Restoration face details comparison. From left to right: ground truth, noise corrupted LR images, results from SGEN, results from SGEN-MSE, results from SRCNN, results from SRResNet, results from SRGAN, results from RED-Net, results from URDGN.}
  \centering
\label{fig:fig_detail}
\end{figure}
We compare the performance of SGEN with five state-of-the-art image restoration networks: SRCNN \cite{dong2014learning}, SRResNet\cite{ledig2017photo}, SRGAN\cite{ledig2017photo}, RED-Net \cite{mao2016image} and URDGN \cite{yu2016ultra}. All the networks are retrained on the same multi-scale and noisy face dataset. The quantitative results are shown in Table \ref{table1} and Table \ref{table2}. The visual restoration samples from one single scale and six scales are shown in Figure \ref{fig:fig6}, Figure \ref{fig:fig_detail} and Figure \ref{fig:fig8}. The quantitative results confirm that our SGEN with only MSE loss achieves state-of-the-art performance in terms of PSNR and SSIM. MOS results from subjective evaluation also suggest that SGEN with adversarial training produces better perceptual quality than other algorithms. From Figure \ref{fig:fig6}, Figure \ref{fig:fig_detail} and Figure \ref{fig:fig8}, we can see that results from SGEN are more visually clear and contain more face details, such as more accurate shape of eyes and mouth in Figure \ref{fig:fig_detail}. Compared with non-ensemble structure SRCNN and URGAN, our SGEN shows better ability to handle multi-scale face restoration. Although the RED-Net, SRResNet and SRGAN try to restore image details with skip-connections that can be viewed as one kind of ensemble structure \cite{veit2016residual}, the networks can not restore faces as good as SGEN due to lack of sequential ensemble method that can sequentially choose to extract high level information from corrupted input and restore more low level details.
\subsection{Gate Layer Pattern Visualization}\label{exp_vis}
\begin{figure}
  \centerline{\includegraphics[width=8.5cm]{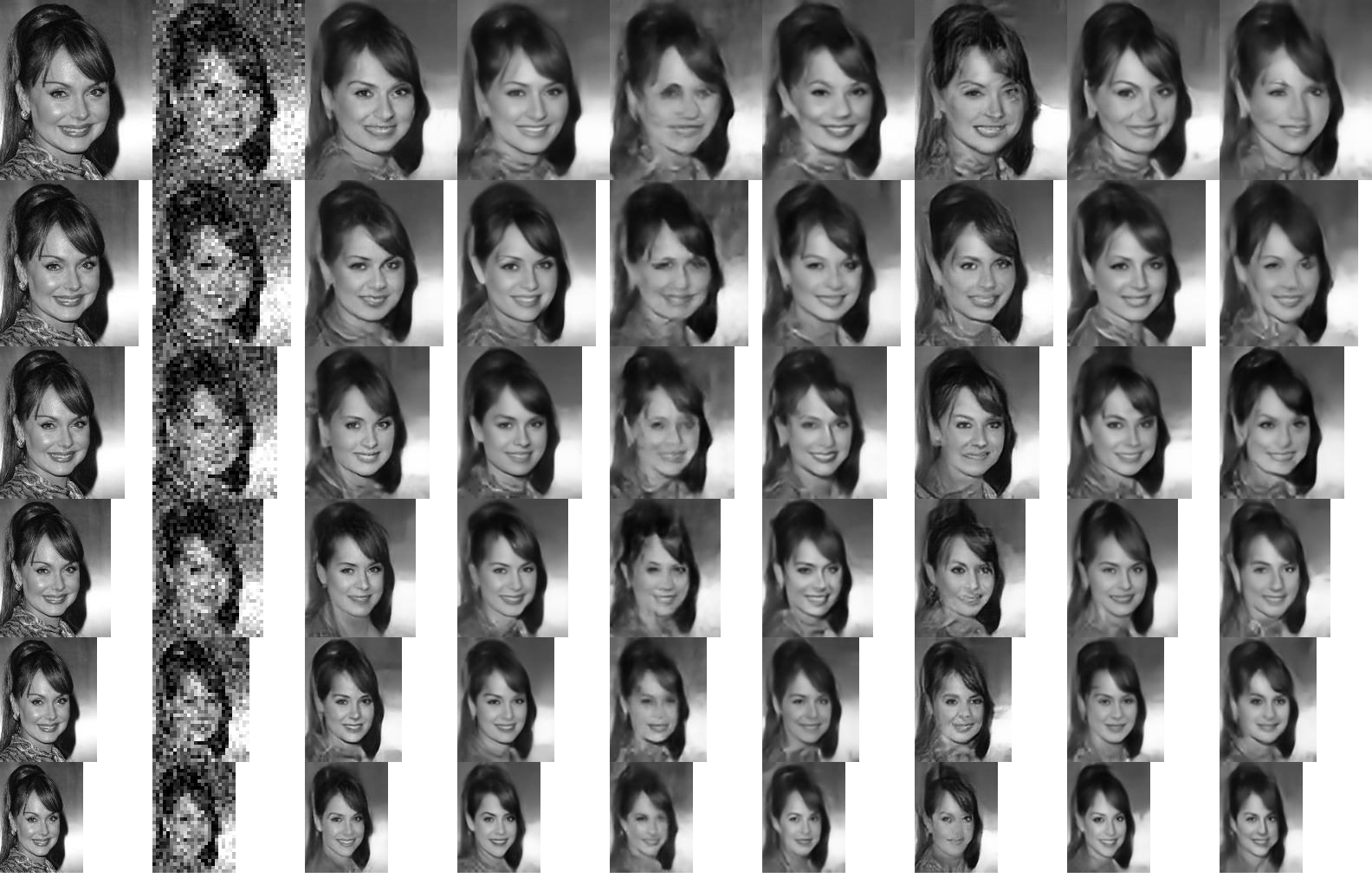}}
  \caption{Restoration samples from the six different scales. From left to right: ground truth, noise corrupted LR images, results from SGEN, results from SGEN-MSE, results from SRCNN, results from SRResNet, results from SRGAN, results from RED-Net, results from URDGN.}
  \centering
\label{fig:fig8}
\end{figure}
\begin{figure}
  \centerline{\includegraphics[width=9.5cm]{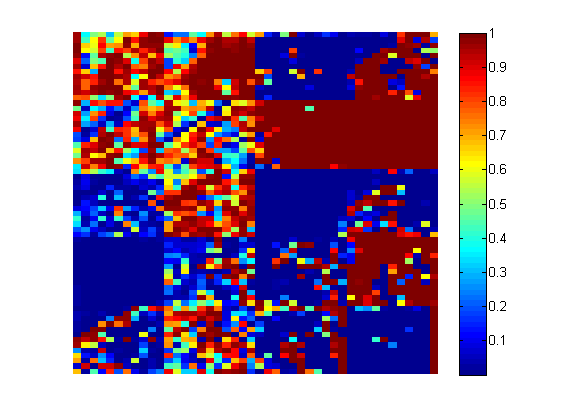}}
  \caption{Visualization of SGU gates in the encoder stage. Columns from left to right: passive and active gates in SGU between $\hat{X}_{1}$ and $X_{2}$, passive and active gates in SGU between $\hat{X}_{2}$ and $X_{3}$.}
  \centering
\label{fig:encoder_gate_layer}
\end{figure}
\begin{figure}
  \centerline{\includegraphics[width=9.5cm]{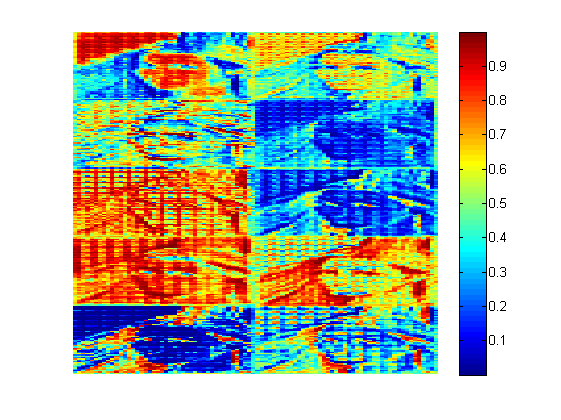}}
  \caption{Visualization of SGU gates between $\hat{Y}_{2}$ and $Y_{3}$. Columns from left to right: passive and active gates in SGU. From top to down, we observe that different feature maps have different information selection.}
  \centering
\label{fig:decoder_gate_layer1}
\end{figure}
\begin{figure}
  \centerline{\includegraphics[width=9.5cm]{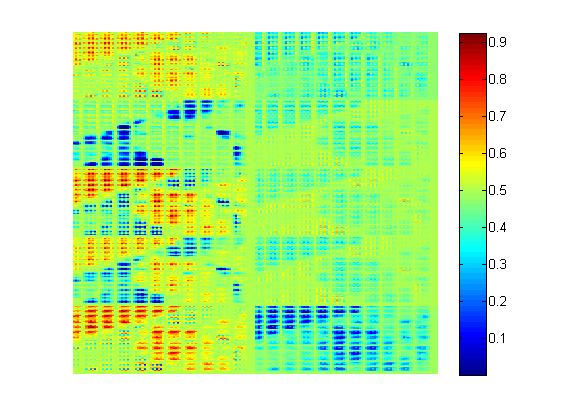}}
  \caption{Visualization of SGU gates between $\hat{Y}_{1}$ and $Y_{2}$. Columns from left to right: passive and active gates in SGU. From top to down, we observe that different feature maps have different information selection.}
  \centering
\label{fig:decoder_gate_layer2}
\end{figure}

In the Sequential Gating Unit (SGU), the gate layers after sigmoid activation decide which information to throw away from $x_a$ and $x_p$ respectively. In the encoder stage, the gates in SGU make selections from high-level information. Thus highly abstract patterns can be observed from the gate layers as shown in Figure~\ref{fig:encoder_gate_layer}. In the decoder stage, the gating operations are implemented after deconvolution, so the information selection is about to restore low details. In the gates between $\hat{Y}_{2}$ and $Y_{3}$ ($N=3$), as shown in Figure~\ref{fig:decoder_gate_layer1}, face patterns are very obvious, which means in our proposed scheme SGUs learn to make selections based on face information. Other than that, different SGUs focus on different landmarks. Because of this, the generated face contains more details. In the gates between $\hat{Y}_{1}$ and $Y_{2}$, as shown in Figure~\ref{fig:decoder_gate_layer2}, the gates contains less visible face details because gates between $\hat{Y}_{1}$ and $Y_{2}$ combine higher level information than ones between $\hat{Y}_{2}$ and $Y_{3}$. Difference can only be told among face, hair and background. We also observe one more interesting pattern that the gate layers for the active input and passive input are complementary to each other. When the active gate selects much from the active input, the values in the passive gate are usually very small, and vice versa. This phenomenon proves that there are redundancy between different scales and selection operation is very important.
\subsection{Computational Time}\label{4.6}
\begin{table*}[]
\centering
\caption{Average computational cost time of different methods.}
\label{avg_time_methods}
\begin{tabular}{|c|c|c|c|c|c|c|}
\hline
              & \textbf{SRCNN} & \textbf{SRResNet} & \textbf{SRGAN} & \textbf{RED-Net} & \textbf{URDGN} & \textbf{SGEN}\\ \hline
\textbf{Training Time} & 10h          & 20h          & 23h          & 15h          & 20h           & 26h \\ \hline
\textbf{Testing Time} & 30s          & 55s          & 55s          & 50s          & 35s           & 64s \\ \hline
\end{tabular}
\end{table*}

We train all the networks on one NVIDIA K80 GPU, and the average computational cost of different methods is shown in Table \ref{avg_time_methods}. The training time of the SGEN ($26$ GPU-hours) is little slower than the structure with the same loss terms but without ensemble, such as URDGN ($20$ GPU-hours). The sequential ensemble structure converges little slower than the single autoencoder structure is not surprising, as the SGEN learns multi-scale features and possesses a larger feature capacity. Considering the fact that our network can converge slightly more than one GPU-day and costs similar testings time as other methods, we conclude that our network structure is a relatively light structure and the convergence is not a problem for our network. In addition, our model only takes about $0.0032\text{s} = 64\text{s}/20000$ on average for each test image, which is well suited for real-time image processing.
\begin{figure}
\centerline{\includegraphics[width=10.5cm]{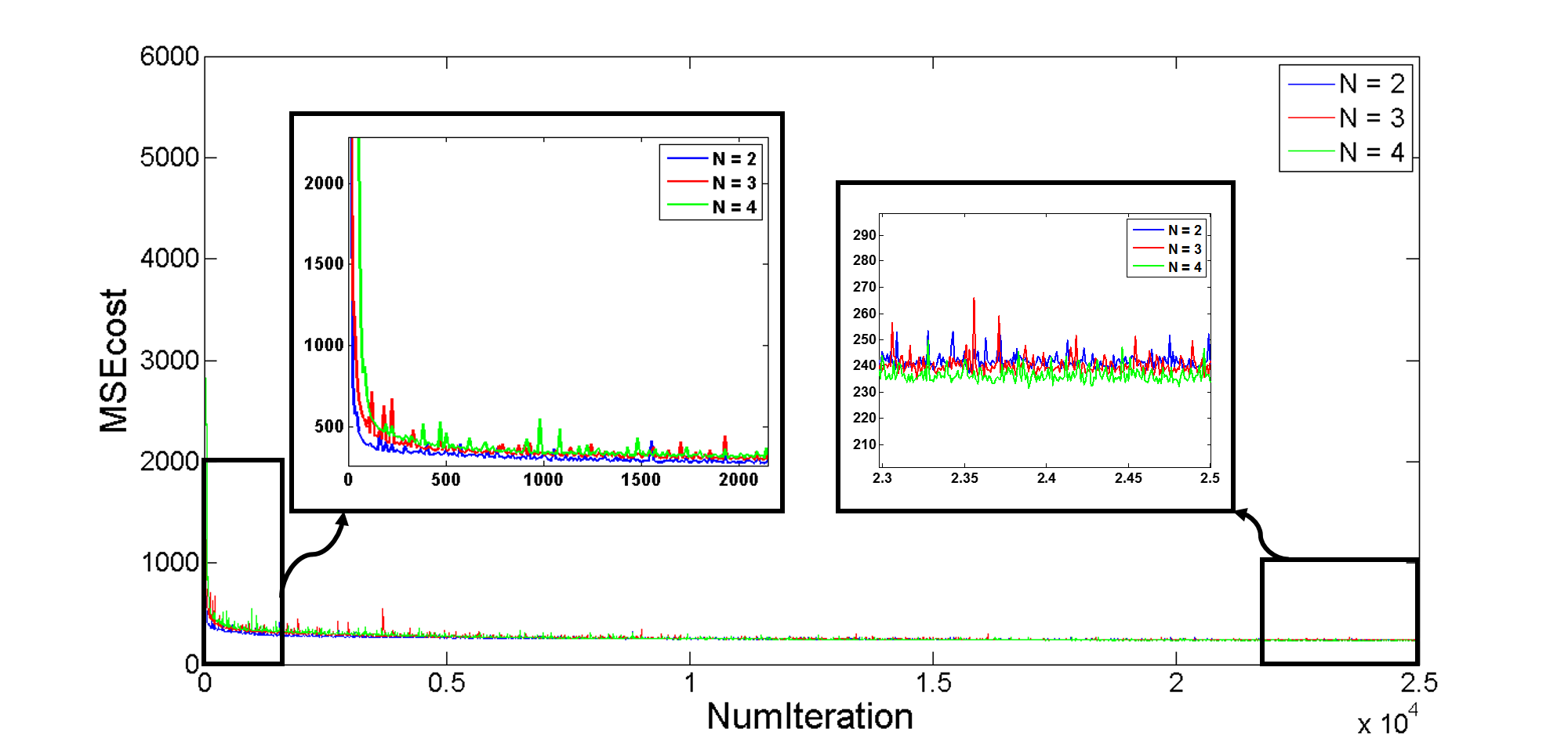}}
\caption{Testing loss while training of SGEN with different $N$.}
\label{fig:loss}
\end{figure}
\subsection{Influence of $N$, AWGN Standard Deviation $\delta$ and Noise Type}\label{4.7}
In addition to SGEN with $N=3$, we train other two SGENs to explore the influence of main parameter $N$ of SGEN. The average PSNR and SSIM performance on test set, and computational cost time of SGEN with different $N$ are shown in Table \ref{table5} and Figure \ref{fig:loss}. Without losing the universality, only MSE loss for SGEN is used for this comparison. It is obvious that the performance of SGEN improves with the increase of $N$. SGEN with $N=4$ achieves performance gain about $0.86\%$ in terms of PSNR against SGEN with $N=3$. However, $35\%$ extra computational cost will also be added by SGEN with $N=4$ compared with $N=3$. Also, as shown in Figure \ref{fig:loss}, the convergence curve of SGEN with $N=3$ behaves much more similarly to $N=4$ than $N=2$, which decreases slower in
the initial stage but converges to a smaller value. Therefore, we choose $N=3$ in our paper to achieve a trade-off between computational cost and performance.

 To further investigate the performance of SGEN on different AWGN noise level and different noise type, we independently train SGEN on images corrupted by different AEGN noise  standard deviation $\delta$ and additive noise sampled from uniform distribution ranging from $0$ to $30$, and the results are shown in Table \ref{table_delta}. We can observe that our SGEN shows robust performance on multi-scale images with different $\delta$ and noise type, which demonstrates the flexibility of SGEN when being applied on a wide range of real scenes.
\begin{table}[]
\centering
\caption{Average PSNR and SSIM performance and computational cost time of SGEN with different $N$.}
\label{table5}
\begin{tabular}{|c|c|c|c|}
\hline
              & \textit{\textbf{N=2}} & \textit{\textbf{N=3}} & \textit{\textbf{N=4}} \\ \hline
\textbf{PSNR} & 24.13        & 24.28        & 24.49        \\ \hline
\textbf{SSIM} & 0.7289       & 0.7373       & 0.7463       \\ \hline
\textbf{Time} & 20h          & 26h          & 35h          \\ \hline
\end{tabular}
\end{table}
\begin{table}[]
\centering
\caption{Aveage PSNR performance of SGEN on images with different AWGN noise standard deviation $\delta$ and additive noise sampled from uniform distribution.}
\label{table_delta}
\begin{tabular}{|c|c|c|c|c|c|}
\hline
              & \textit{\textbf{$\delta=20$}} & \textit{\textbf{$\delta=30$}} & \textit{\textbf{$\delta=40$}} & \textit{\textbf{$\delta=50$}} & Uniform \\ \hline
\textbf{Scale $128 \times 96$} & 23.99        & 23.00        & 22.29       & 21.52  & 24.27 \\ \hline
\textbf{Scale $144 \times 112$} & 24.60       & 23.60       & 22.83      & 22.04 & 25.01 \\ \hline
\textbf{Scale $160 \times 128$} & 25.08        & 24.12        & 23.25       & 22.44  & 25.56 \\ \hline
\textbf{Scale $176 \times 144$} & 25.58        & 24.61        & 23.67       & 22.82  & 26.18 \\ \hline
\textbf{Scale $192 \times 160$} & 26.02        & 24.98        & 24.04       & 23.20  & 26.72 \\ \hline
\textbf{Scale $208 \times 176$} & 26.45        & 25.39        & 24.39       & 23.52  & 27.27 \\ \hline
\end{tabular}
\end{table}
\subsection{Testing beyond Training scales}
\begin{figure}
\centerline{\includegraphics[width=8.5cm]{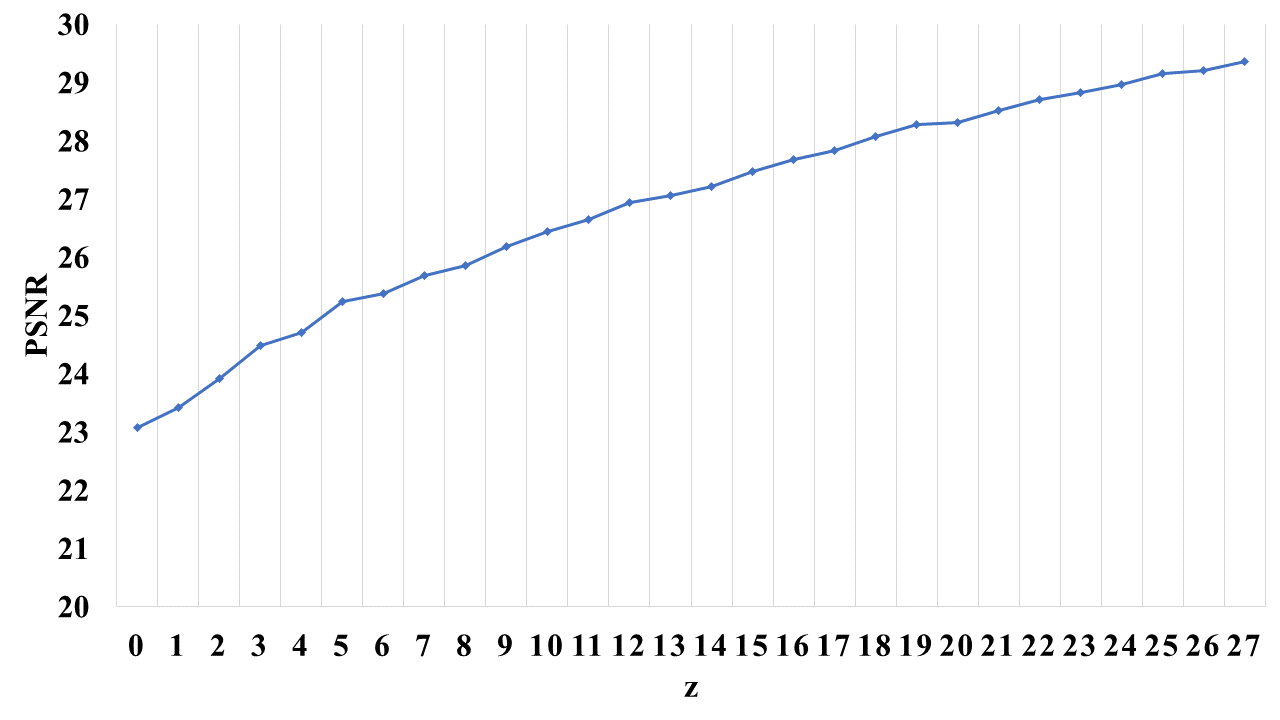}}
\caption{Aveage PSNR performance on images of scale $=(128+16*z) \times (96+16*z)$. }
\label{fig:psnr_scale}
\end{figure}

\begin{figure}
\centerline{\includegraphics[width=8.5cm]{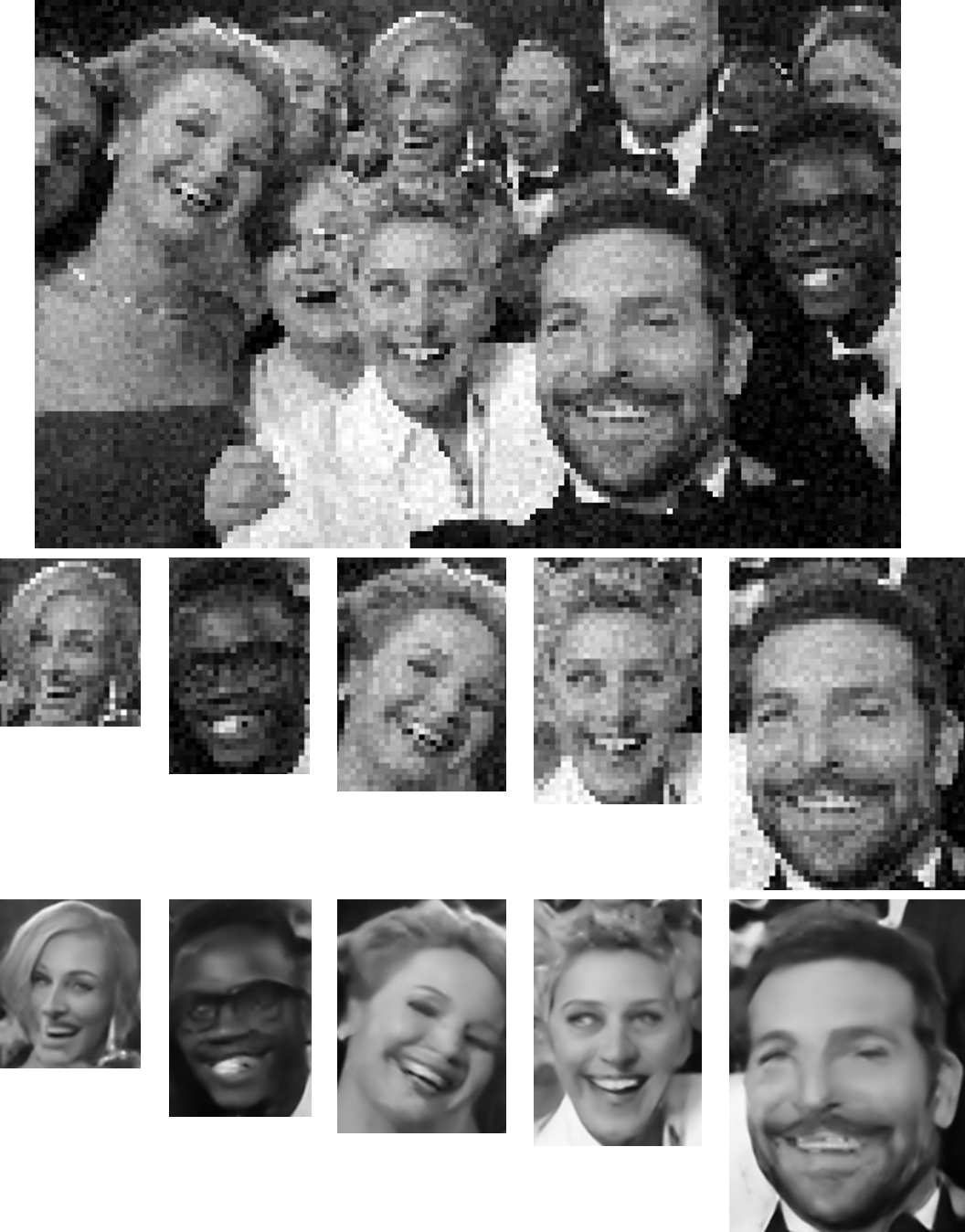}}
\caption{Test on real-world image. Top: LR and noise real-world image. Medium: Extracted LR and noise faces. Bottom: restored results by SGEN.}
\label{fig:blind_test}
\end{figure}

Sizes of face images in real world often vary beyond training scales ($128 \times 96$ to $208 \times 176$). Therefore, we carry out multi-scale face restoration experiments on face image of scale $=(128+16*z )\times (96+16*z), z=0,1,...,27$, with SGEN only trained on scales from $128 \times 96$ to $208 \times 176$. The average PSNR results on a set of test images are shown in Figure \ref{fig:psnr_scale}. In order to further testify the robustness of SGEN, we conduct experiment on LR and noise real-world image as shown in Figure \ref{fig:blind_test}. We can easily found that although SGEN is only trained on faces of limited scales, it shows even better PSNR performance on unseen faces of different scales in Figure \ref{fig:psnr_scale}. The real-world test also confirms that our SGEN can tackle distortions on faces of various orientation, illumination, etc.
\section{Conclusions and Future Work}\label{sec5}
In this paper, we present a SGEN model for multi-scale face restoration from low-resolution and strong noise. We propose to aggregate multi-level base-en/decoders into the SGEN. The SGEN takes en/decoders from different levels as the sequential data. Specifically, the SGEN learns to sequentially extract high level information from low level base-encoders in bottom-up manner and sequentially restore low level details from high level base-decoders in top-down manner. Specially, we propose an elaborate SGU unit that could sequentially combine and select the multi-level information from base-en/decoders. Owing to the sequential ensemble structure, SGEN with MSE loss achieves the state-of-the-art performance of multi-scale face restoration in terms of PSNR and SSIM. Further applying adversarial loss to SGEN training, SGEN achieves the best perceptual quality according to subjective evaluation.

In future work, we would to explore SGEN in following multiple aspects. First, we can apply the proposed model to other image-to-image translation tasks, such as image restoration and style transformation. Second, it is worth exploring SGEN on face video restoration by combining temporal information in the future, which can be utilized for real-time surveillance video analysis. Third, it is also interesting to combine SGEN with face analysis techniques into one end-to-end training scheme for both face restoration and face analysis.

\ifCLASSOPTIONcaptionsoff
  \newpage
\fi



%
{\small
\bibliographystyle{unsrt}
\bibliography{ref}

\begin{thebibliography}{10}

\bibitem{wang2005hallucinating}
Xiaogang Wang and Xiaoou Tang.
\newblock Hallucinating face by eigentransformation.
\newblock {\em IEEE Transactions on Systems, Man, and Cybernetics, Part C
  (Applications and Reviews)}, 35(3):425--434, 2005.

\bibitem{ma2010hallucinating}
Xiang Ma, Junping Zhang, and Chun Qi.
\newblock Hallucinating face by position-patch.
\newblock {\em Pattern Recognition}, 43(6):2224--2236, 2010.

\bibitem{wang2014face}
Zhongyuan Wang, Ruimin Hu, Shizheng Wang, and Junjun Jiang.
\newblock Face hallucination via weighted adaptive sparse regularization.
\newblock {\em IEEE Transactions on Circuits and Systems for video Technology},
  24(5):802--813, 2014.

\bibitem{yu2016ultra}
Xin Yu and Fatih Porikli.
\newblock Ultra-resolving face images by discriminative generative networks.
\newblock In {\em European Conference on Computer Vision}, pages 318--333.
  Springer, 2016.

\bibitem{huang2011fast}
Hua Huang and Ning Wu.
\newblock Fast facial image super-resolution via local linear transformations
  for resource-limited applications.
\newblock {\em IEEE Transactions on Circuits and Systems for Video Technology},
  21(10):1363--1377, 2011.

\bibitem{zeng2018expanding}
Xiao Zeng, Hua Huang, and Chun Qi.
\newblock Expanding training data for facial image super-resolution.
\newblock {\em IEEE transactions on cybernetics}, 48(2):716--729, 2018.

\bibitem{jiang2014noise}
Junjun Jiang, Ruimin Hu, Zhongyuan Wang, and Zhen Han.
\newblock Noise robust face hallucination via locality-constrained
  representation.
\newblock {\em IEEE Transactions on Multimedia}, 16(5):1268--1281, 2014.

\bibitem{jiang2016noise}
Junjun Jiang, Jiayi Ma, Chen Chen, Xinwei Jiang, and Zheng Wang.
\newblock Noise robust face image super-resolution through smooth sparse
  representation.
\newblock {\em IEEE Transactions on Cybernetics}, 2016.

\bibitem{taigman2016unsupervised}
Yaniv Taigman, Adam Polyak, and Lior Wolf.
\newblock Unsupervised cross-domain image generation.
\newblock {\em arXiv preprint arXiv:1611.02200}, 2016.

\bibitem{yoo2016pixel}
Donggeun Yoo, Namil Kim, Sunggyun Park, Anthony~S Paek, and In~So Kweon.
\newblock Pixel-level domain transfer.
\newblock In {\em European Conference on Computer Vision}, pages 517--532.
  Springer, 2016.

\bibitem{johnson2016perceptual}
Justin Johnson, Alexandre Alahi, and Li~Fei-Fei.
\newblock Perceptual losses for real-time style transfer and super-resolution.
\newblock In {\em European Conference on Computer Vision}, pages 694--711.
  Springer, 2016.

\bibitem{hinton2006reducing}
Geoffrey~E Hinton and Ruslan~R Salakhutdinov.
\newblock Reducing the dimensionality of data with neural networks.
\newblock {\em science}, 313(5786):504--507, 2006.

\bibitem{kuncheva2004combining}
Ludmila~I Kuncheva.
\newblock {\em Combining pattern classifiers: methods and algorithms}.
\newblock John Wiley \& Sons, 2004.

\bibitem{polikar2006ensemble}
Robi Polikar.
\newblock Ensemble based systems in decision making.
\newblock {\em IEEE Circuits and systems magazine}, 6(3):21--45, 2006.

\bibitem{hochreiter1997long}
Sepp Hochreiter and J{\"u}rgen Schmidhuber.
\newblock Long short-term memory.
\newblock {\em Neural computation}, 9(8):1735--1780, 1997.

\bibitem{sundermeyer2012lstm}
Martin Sundermeyer, Ralf Schl{\"u}ter, and Hermann Ney.
\newblock Lstm neural networks for language modeling.
\newblock In {\em Interspeech}, pages 194--197, 2012.

\bibitem{sutskever2014sequence}
Ilya Sutskever, Oriol Vinyals, and Quoc~V Le.
\newblock Sequence to sequence learning with neural networks.
\newblock In {\em Advances in neural information processing systems}, pages
  3104--3112, 2014.

\bibitem{goodfellow2014generative}
Ian Goodfellow, Jean Pouget-Abadie, Mehdi Mirza, Bing Xu, David Warde-Farley,
  Sherjil Ozair, Aaron Courville, and Yoshua Bengio.
\newblock Generative adversarial nets.
\newblock In {\em Advances in neural information processing systems}, pages
  2672--2680, 2014.

\bibitem{denton2015deep}
Emily~L Denton, Soumith Chintala, Rob Fergus, et~al.
\newblock Deep generative image models using a laplacian pyramid of adversarial
  networks.
\newblock In {\em Advances in neural information processing systems}, pages
  1486--1494, 2015.

\bibitem{radford2015unsupervised}
Alec Radford, Luke Metz, and Soumith Chintala.
\newblock Unsupervised representation learning with deep convolutional
  generative adversarial networks.
\newblock {\em arXiv preprint arXiv:1511.06434}, 2015.

\bibitem{salimans2016improved}
Tim Salimans, Ian Goodfellow, Wojciech Zaremba, Vicki Cheung, Alec Radford, and
  Xi~Chen.
\newblock Improved techniques for training gans.
\newblock In {\em Advances in Neural Information Processing Systems}, pages
  2226--2234, 2016.

\bibitem{chakrabarti2007super}
Ayan Chakrabarti, AN~Rajagopalan, and Rama Chellappa.
\newblock Super-resolution of face images using kernel pca-based prior.
\newblock {\em IEEE Transactions on Multimedia}, 9(4):888--892, 2007.

\bibitem{zhuang2007hallucinating}
Yueting Zhuang, Jian Zhang, and Fei Wu.
\newblock Hallucinating faces: Lph super-resolution and neighbor reconstruction
  for residue compensation.
\newblock {\em Pattern Recognition}, 40(11):3178--3194, 2007.

\bibitem{huang2010super}
Hua Huang, Huiting He, Xin Fan, and Junping Zhang.
\newblock Super-resolution of human face image using canonical correlation
  analysis.
\newblock {\em Pattern Recognition}, 43(7):2532--2543, 2010.

\bibitem{yang2010image}
Jianchao Yang, John Wright, Thomas~S Huang, and Yi~Ma.
\newblock Image super-resolution via sparse representation.
\newblock {\em IEEE transactions on image processing}, 19(11):2861--2873, 2010.

\bibitem{lecun1989backpropagation}
Yann LeCun, Bernhard Boser, John~S Denker, Donnie Henderson, Richard~E Howard,
  Wayne Hubbard, and Lawrence~D Jackel.
\newblock Backpropagation applied to handwritten zip code recognition.
\newblock {\em Neural computation}, 1(4):541--551, 1989.

\bibitem{he2016deep}
Kaiming He, Xiangyu Zhang, Shaoqing Ren, and Jian Sun.
\newblock Deep residual learning for image recognition.
\newblock In {\em Proceedings of the IEEE conference on computer vision and
  pattern recognition}, pages 770--778, 2016.

\bibitem{ren2015faster}
Shaoqing Ren, Kaiming He, Ross Girshick, and Jian Sun.
\newblock Faster r-cnn: Towards real-time object detection with region proposal
  networks.
\newblock In {\em Advances in neural information processing systems}, pages
  91--99, 2015.

\bibitem{parkhi2015deep}
Omkar~M Parkhi, Andrea Vedaldi, Andrew Zisserman, et~al.
\newblock Deep face recognition.
\newblock In {\em BMVC}, volume~1, page~6, 2015.

\bibitem{long2015fully}
Jonathan Long, Evan Shelhamer, and Trevor Darrell.
\newblock Fully convolutional networks for semantic segmentation.
\newblock In {\em Proceedings of the IEEE Conference on Computer Vision and
  Pattern Recognition}, pages 3431--3440, 2015.

\bibitem{dong2014learning}
Chao Dong, Chen~Change Loy, Kaiming He, and Xiaoou Tang.
\newblock Learning a deep convolutional network for image super-resolution.
\newblock In {\em European Conference on Computer Vision}, pages 184--199.
  Springer, 2014.

\bibitem{veit2016residual}
Andreas Veit, Michael~J Wilber, and Serge Belongie.
\newblock Residual networks behave like ensembles of relatively shallow
  networks.
\newblock In {\em Advances in Neural Information Processing Systems}, pages
  550--558, 2016.

\bibitem{ledig2017photo}
Christian Ledig, Lucas Theis, Ferenc Husz{\'a}r, Jose Caballero, Andrew
  Cunningham, Alejandro Acosta, Andrew~P Aitken, Alykhan Tejani, Johannes Totz,
  Zehan Wang, et~al.
\newblock Photo-realistic single image super-resolution using a generative
  adversarial network.
\newblock In {\em CVPR}, volume~2, page~4, 2017.

\bibitem{mao2016image}
Xiaojiao Mao, Chunhua Shen, and Yu-Bin Yang.
\newblock Image restoration using very deep convolutional encoder-decoder
  networks with symmetric skip connections.
\newblock In {\em Advances in Neural Information Processing Systems}, pages
  2802--2810, 2016.

\bibitem{nair2010rectified}
Vinod Nair and Geoffrey~E Hinton.
\newblock Rectified linear units improve restricted boltzmann machines.
\newblock In {\em Proc. ICML}, pages 807--814, 2010.

\bibitem{maas2013rectifier}
Andrew~L Maas, Awni~Y Hannun, and Andrew~Y Ng.
\newblock Rectifier nonlinearities improve neural network acoustic models.
\newblock In {\em Proc. ICML}, volume~30, 2013.

\bibitem{lin2013network}
Min Lin, Qiang Chen, and Shuicheng Yan.
\newblock Network in network.
\newblock {\em arXiv preprint arXiv:1312.4400}, 2013.

\bibitem{kingma2014adam}
Diederik Kingma and Jimmy Ba.
\newblock Adam: A method for stochastic optimization.
\newblock {\em arXiv preprint arXiv:1412.6980}, 2014.

\bibitem{nesterov1983method}
Yurii Nesterov.
\newblock A method of solving a convex programming problem with convergence
  rate o (1/k2).
\newblock In {\em Soviet Mathematics Doklady}, volume~27, pages 372--376, 1983.

\bibitem{liu2015faceattributes}
Ziwei Liu, Ping Luo, Xiaogang Wang, and Xiaoou Tang.
\newblock Deep learning face attributes in the wild.
\newblock In {\em Proceedings of International Conference on Computer Vision
  (ICCV)}, 2015.

\bibitem{wang2004image}
Zhou Wang, Alan~C Bovik, Hamid~R Sheikh, and Eero~P Simoncelli.
\newblock Image quality assessment: from error visibility to structural
  similarity.
\newblock {\em IEEE transactions on image processing}, 13(4):600--612, 2004.

\end{thebibliography}
}

%





\end{document}